\documentclass{vgtc}                          

\ifpdf
  \pdfoutput=1\relax                   
  \usepackage{graphicx}                
  \DeclareGraphicsExtensions{.pdf,.png,.jpg,.jpeg} 
\else
  \usepackage{graphicx}                
  \DeclareGraphicsExtensions{.eps}     
\fi%

\graphicspath{{figures/}{pictures/}{images/}{./}} 

\usepackage{microtype}                 
\PassOptionsToPackage{warn}{textcomp}  
\usepackage{textcomp}                  
\usepackage{mathptmx}                  
\usepackage{times}                     
\usepackage{cite}                      
\usepackage{tabu}                      
\usepackage{booktabs}                  

\usepackage{amsmath}
\usepackage{amssymb}
\usepackage{amsfonts}

\onlineid{1067}

\vgtccategory{Research}

\vgtcinsertpkg

\title{ Video synthesis of human upper body with realistic face}

\author{Zhaoxiang Liu\thanks{e-mail: robin.liu@cloudminds.com}\\ %
        \scriptsize Clouminds %
\and Huan Hu\thanks{e-mail: hans.hu@cloudminds.com}\\ %
     \scriptsize Clouminds %
\and Zipeng Wang\thanks{e-mail: kohou.wang@cloudminds.com}\\ %
     \scriptsize Clouminds %
\and Kai Wang\thanks{e-mail: kai.wang@cloudminds.com}\\ %
     \scriptsize Clouminds %
\and Jinqiang Bai\thanks{e-mail: baijinqiang@buaa.edu.cn}\\ %
     \scriptsize Beihang University %
\and Shiguo Lian\thanks{e-mail: sg\_lian@163.com}\\ %
     \scriptsize Clouminds %
     }

\abstract{This paper presents a generative adversarial learning-based human upper body video synthesis approach to generate an upper body video of target person that is consistent with the body motion, face expression, and pose of the person in source video. We use upper body keypoints, facial action units and poses as intermediate representations between source video and target video. Instead of directly transferring the source video to the target video, we firstly map the source person's facial action units and poses into the target person's facial landmarks, then combine the normalized upper body keypoints and generated facial landmarks with spatio-temporal smoothing to generate the corresponding target video's image. Experimental results demonstrated the effectiveness of our method.%
} 

\CCScatlist{
  \CCScatTwelve{Computing methodologies}{Computer graphics}{Graphics systems and interfaces}{Mixed/augmented reality};
  \CCScatTwelve{Computing methodologies}{Computer graphics}{Image manipulation}{Image-based rendering}
}

\begin{document}


\firstsection{Introduction}

\maketitle

Recently, with the dramatic advances of Generative Adversarial Network (GAN)~\cite{isola2017image-to-image, wang2018high-resolution, zhu2017unpaired, wang2018video-to-video}, GAN-based human body information transfer, such as body motion transfer~\cite{chan2018everybody} and face transfer~\cite{wu2018reenactgan, bansal2018recycle-gan}, which avoids expensive 3D motion capture, has gained much attention. However, these works mainly focus on either holistic transfer or local transfer. Consequently, facial details are not realistic and body motions will be lost. This limits their further applications. Therefore, transferring human motion while keeping realistic facial expressions becomes more and more essential and promising.

In this work, we aim to transfer an upper body video of a source person to a target person. Given an upper body video of a source person and another video of a target person, our goal is to generate an upper body video of the target person that is consistent with the body motion, face expression, and pose of the person in source video. Although it seems intuitive to directly combine body motion transfer with face transfer to realize this goal, this scheme remains impractical due to the difficulty of seamlessly fusing the resulting images of the two separate GANs.

\begin{figure*}
\setlength{\abovecaptionskip}{-0.5cm}
\setlength{\belowcaptionskip}{-0.5cm}
\begin{center}
    \includegraphics[width=0.8\linewidth]{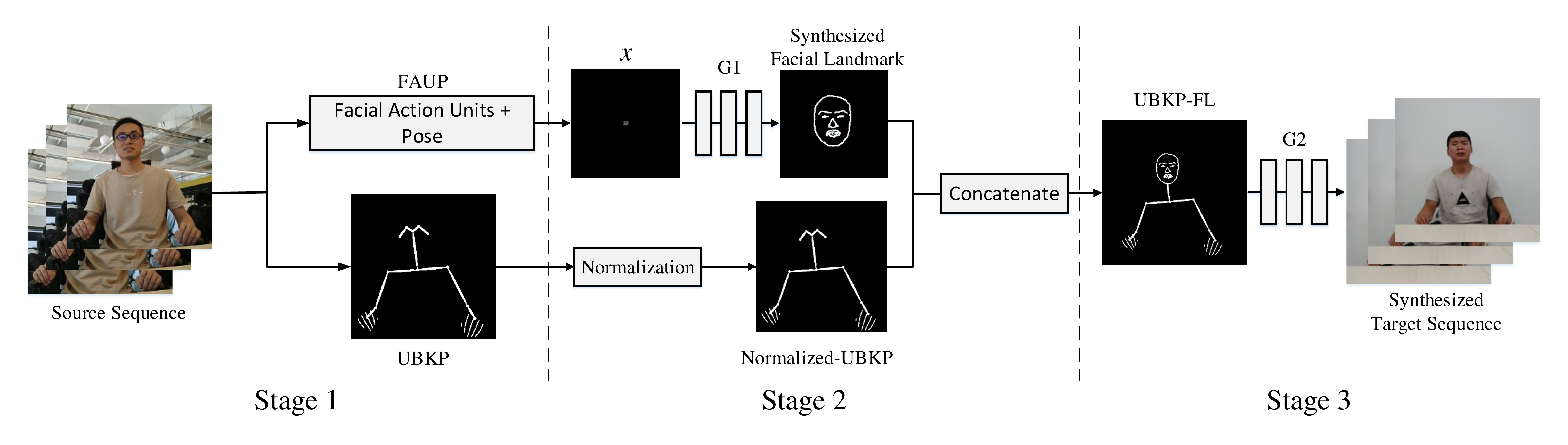}
\end{center}
   \caption{The pipeline of transferring motion from source to target. Stage1: UBKP estimation and FAUP extraction from source video. Stage 2: generation of target's UBKP and facial landmark. Stage 3: synthesis of target video.}
\label{fig1}
\end{figure*}

Upper body keypoints (UBKP) is an ideal intermediate representation for motion transfer between two subjects, and it has been used in body motion transfer~\cite{chan2018everybody}. Nonetheless, UBKP employs only five sparse keypoints to represent a face, and it is thus not enough for transferring facial expressions and local movements. Facial landmark is a popular intermediate representation for realistic face synthesis \cite{wang2018high-resolution}. An intuitive idea is to combine the facial landmarks and UBKP for body motion transfer with realistic face reenactment. However, as the facial landmark is highly relevant to the identity, it is extremely difficult to normalize the source's facial landmarks into target's using linear transformation as~\cite{chan2018everybody} did for body keypoints normalization.

To address this issue, we propose to combine UBKP, facial action units and pose(FAUP) to realize body motion transfer, to utilize FAUP instead of facial landmarks to enhance the details of the synthesized face. We firstly map the source's FAUP into target's facial landmarks, and then translate the landmarks into target's face image. Compared to facial landmarks, FAUP have the advantages of being independent of identity and containing no spatial coordinate information. It thus can provide an abstract description of facial expressions and local movements without complicated normalization. The proposed pipeline is able to transfer both the human upper body motion and face information at the same time.


\section{Related work}

Video synthesis of human can mostly be implemented by using face reenactment and body reenactment, both of
 which have been investigated for years. Here we mainly review GAN-based methods.

{\bfseries Face reenactment.} With the proposal of Generative Adversarial Network (GAN) \cite{goodfellow2014generative}, image generation without explicit feature analysis has become possible. The Pix2Pix\cite{isola2017image-to-image} has shown the potential of applying an image to image translation technique to face reenactment. By adding a cycle consistency loss, CycleGAN\cite{zhu2017unpaired} allows the cross-transfer of two domains, which can be used to the transfer pose and expressions between the two faces in the source and target videos. By making use of CycleGAN, the works \cite{ bansal2018recycle-gan, jin2017cyclegan, xu2017face} have been able to realize one-to-one mapping of facial expressions in two videos. To allow the mapping from multiple faces in the source video to a face in the target video, ReenactGAN \cite{wu2018reenactgan} mapped the faces into a geometric latent space and decoding it to specific persons. The boundary latent space helps to ensure the consistency of facial actions and is rather robust to extreme poses.

{\bfseries Body reenactment.} Similar to face reenactment, GAN \cite{goodfellow2014generative} has also been widely used to generate new body poses. The works \cite{esser2018a, joo2018generating, Ma2017Pose} have been able to generate new poses in static images by making use of GAN. \cite{tulyakov2018mocogan} proposed to learn human movements in videos and generate new motions with an unsupervised adversarial network. \cite{baddar2017dynamics} improved it to allow the transfer of motions or facial expressions in a source video to a person in a static image. \cite{balakrishnan2018synthesizing} presented a network to synthesize temporally coherent new poses in a video, and \cite{villegas2017learning} further allows the prediction of subsequent body movement. \cite{chan2018everybody} proposes to transfer the body movement from a source video to a target using pix2pixHD \cite{wang2018high-resolution}.

Different from above works, We propose a novel pipeline instead of a simple combination of face reenactment and body reenactment to implement human upper body motion transfer with realistic face synthesis.

\section{Method overview}

Our goal is to generate an upper body video of the target person that is consistent with the motion, expression, and pose of the person in source video. To do that, we use upper body keypoints, facial action units and face poses as compact media to capture body and facial's geometric variances. The proposed pipeline can be divided into three stages(as shown in Fig.~\ref{fig1}): UBKP estimation and FAUP extraction from source video, generation of target's UBKP and facial landmark, and synthesis of target video.


In stage 1, a pre-trained pose detector~\cite{cao2017realtime,simon2017hand} is employed to estimate the upper body joint coordinates from source video. The stage 1 of Fig.~\ref{fig1} shows a representation of the resulting pose stick figure by plotting the keypoints and drawing lines between connected joints. Meanwhile, the sequence of FAUP of the source video is estimated using a popular framework \cite{baltruvsaitis2016openface}. In stage 2, the target's UBKP and facial landmark are generated separately. For the generation of the UBKP, it is necessary to transform that of the source person so that they appear in accordance with the target person's body shape. The global pose normalization method presented in~\cite{chan2018everybody} is then employed to linearly transform UBKP between the source and the target. For the generation of target's facial landmark, a GAN-based synthesizer is utilized to build the mapping from source's FAUP to target's facial landmark. Next the target's UBKP and facial landmark are concatenated together into one image to serve as the input of stage 3. In stage 3, another GAN-based synthesizer is employed to generate the target's upper body video with realistic face synthesis from UBKP and facial landmark.

\section{Generation of target's facial landmark}

\begin{figure}[htbp]
	\begin{center}
		\includegraphics[width=\linewidth]{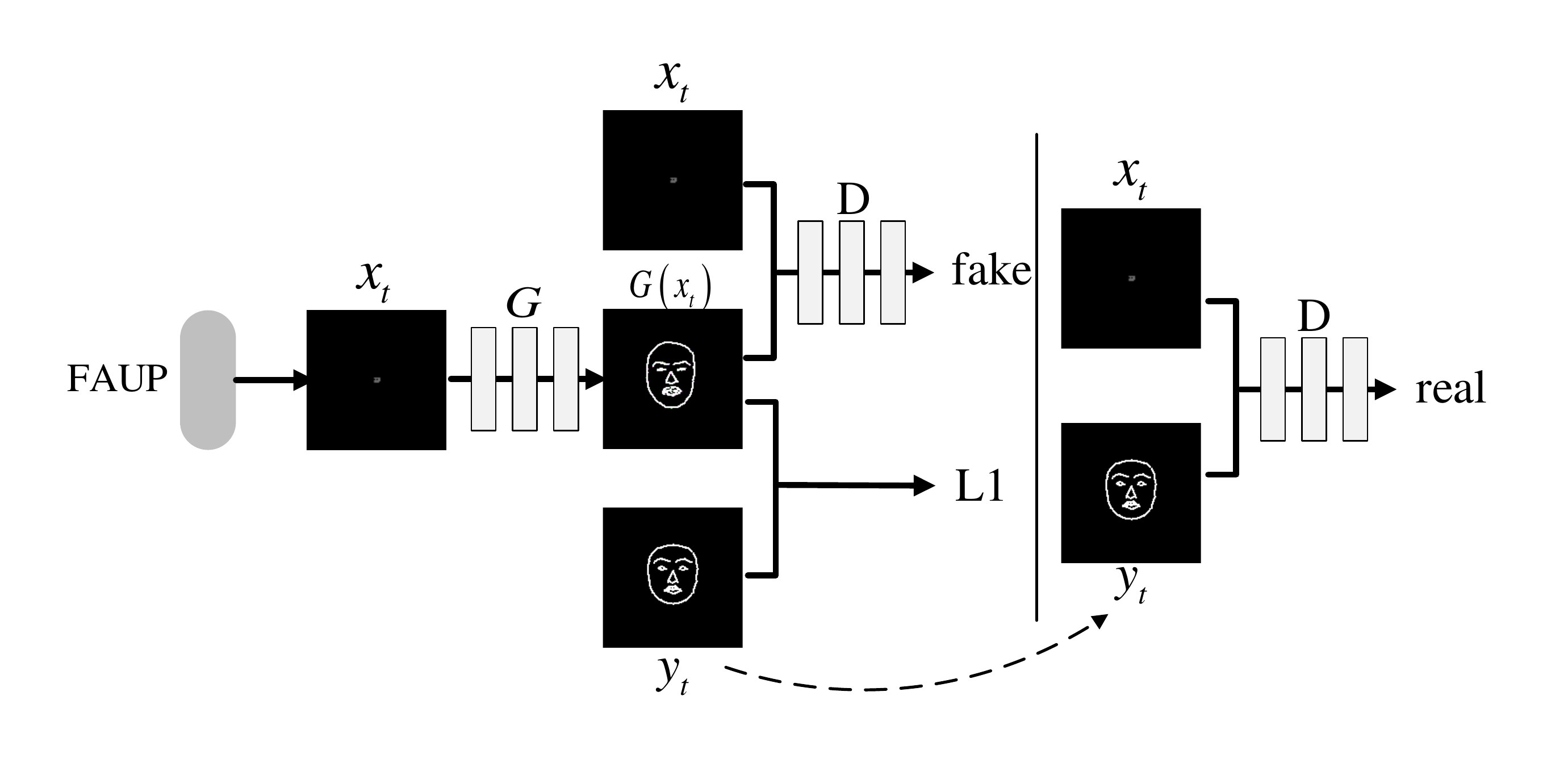}
	\end{center}
	\caption{We first map FAUP to image $x_t$. Then learn the mapping $G$ to generate $G(x_t)$. $L1$ reconstruction loss is used to reduce the gap between $G(x_t)$ and ground truth $y_t$.The discriminator $D$ attempts to differentiate the real frame from the fake image.}
	\label{fig3}
\end{figure}

To use FAUP to synthesize the facial landmark of the target person, we adopt the state-of-the-art generative adversarial network pix2pixHD~\cite{wang2018high-resolution} to complete this transfer. For our task, the objective of the generator $G$ is to translate FAUP maps to realistic-looking facial landmark images, while the discriminator $D$ is used to distinguish real images from the translated ones. During the training phase, target's FAUP and facial landmark are employed to train generator $G$ and discriminator $D$. During the transferring phase, we directly use source's FAUP to synthesize target's facial landmark. Fig.~\ref{fig3} shows the training setup. Since the extracted FAUP does not have coordinate information like UBKP, we fill the FAUP vector into the center of an empty image and then feed it into our network for training, in order to facilitate loading data in the training phase. This also allows our network to have spatial coordinate constraints during training, which can accelerate the convergence. The objective function is given by:

\begin{equation}
\begin{aligned}
L_{GAN}(G,D_k)=\mathbb{E}_{(x,y)}[logD_k(x,y)]\\
     + \mathbb{E}_{x}[log(1-D_k(x,G(x)))] \qquad¡¡k=1,2,3.
\label{equation1}
\end{aligned}
\end{equation}

Finally, we employ a combined loss which is widely used in various  data generation tasks:
\begin{equation}
\begin{aligned}
\begin{split}
    L={L}_{GAN}(G,D)+\lambda *{L}_{L_1}(G(x),y)
\end{split}
\end{aligned}
\label{equation2}
\end{equation}
where the first term is the adversarial loss in Eq.~\ref{equation1}, the second term is an $L1$ reconstruction loss that measures the pixel-level deviation between the synthesized facial landmark image $G(x)$ and ground truth $y$, and $\lambda$ controls the importance of the two terms.

\section{Synthesis of target's upper body video }

We also use the pix2pixHD~\cite{wang2018high-resolution} framework as our backbone network which takes concatenated UBKP and facial landmark(UBKP-FL) image as input to generate target's image. During the training phase, the target's UBKP-FL images and video images are employed to train the framework. In the transferring phase, we directly use concatenated UBKP-FL images generated from the source video to synthesize the target video.

\begin{figure}
	\begin{center}
		\includegraphics[width=\linewidth]{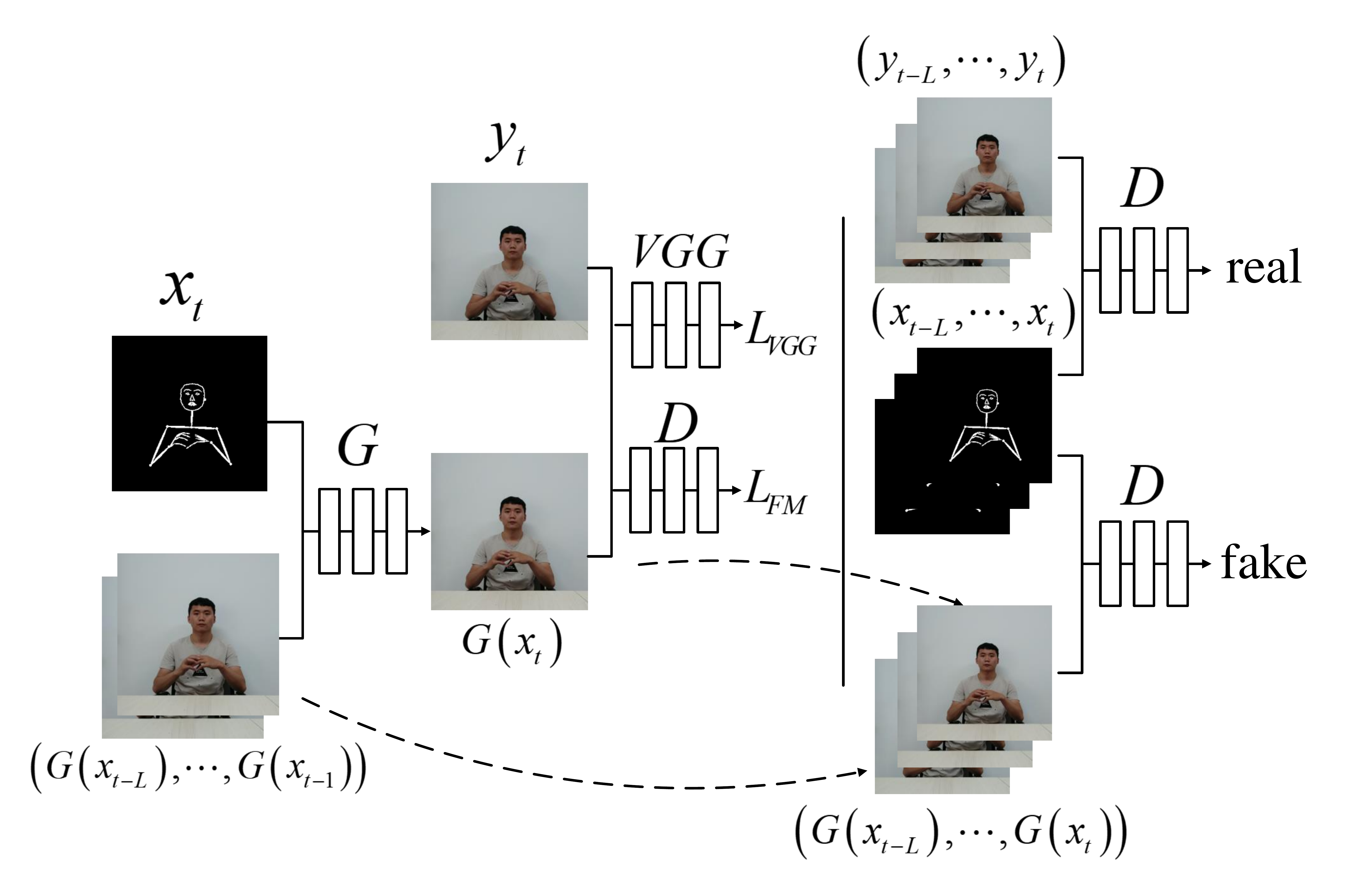}
	\end{center}
	\caption{We use pair $(x_t,(G(x_{t-L}),...,G(x_{t-1})))$ to learn the mapping $G$ to generate $G(x_t)$. VGG perceptual reconstruction loss and feature matching loss $L_{FM}$ are used to reduce the gap between $G(x_t)$ and ground truth $y_t$.The discriminator $D$ attempts to differentiate the real temporally coherent sequence from the fake sequence.}
	\label{fig5}
\end{figure}

However, without considering the temporal correlation, directly applying pix2pixHD to convert the source video into target video will result in temporally incoherent videos of low visual quality. Inspired by \cite{chan2018everybody}, we modify the adversarial training setup of pix2pixHD~\cite{wang2018high-resolution} to produce temporally coherent video frames (as shown in Fig.~\ref{fig5}). We similarly assume the video frames are generated sequentially, and the generation of the current frame only depends on the current source frame, past $L$ source frames and past $L$ generated frames. We predict $L$ consecutive frames where the first output $G(x_{t-L})$ is conditioned on its corresponding UBKP-FL image $x_{t-L}$ and a zero image $z$. The last output $G(x_t)$ is conditioned on its corresponding UBKP-FL figure $x_t$ and the previous $L-1$ frames output $(G(x_{t-L}),...,G(x_{t-1}))$ . The temporal smoothing loss $L_{tS}(G,D)$ is as following:

\begin{equation}
\begin{aligned}
    L_{tS}(G,D)&=\mathbb{E}_{(x,y)}[logD((x_{x-L},...,x_t),(y_{t-L},...,y_t))]\\
     &+\mathbb{E}_{x}[log(1-D((x_{t-L},...,x_t),\\
     &(G(x_{t-L}),...,G(x_t))))]
\end{aligned}
\label{equation5}
\end{equation}

Finally, our framework is trained with the temporal smoothing loss $L_{tS}(G,D)$, feature matching loss $L_{FM}$~\cite{wang2018high-resolution}, and perceptual reconstruction loss $L_{VGG}$~\cite{wang2018high-resolution}, as:
\begin{equation}
\begin{aligned}
    \mathop {\min }\limits_G(\mathop{\max} \limits_{D_1,D_2,D_3}\sum _{k=1,2,3}L_{tS}(G,D_k)
&+\alpha \sum _{k=1,2,3}L_{FM}(G,D_k)\\
&+\beta \sum_{i=0}^{L}L_{VGG}(G(x_{t-i}),y_{t-i}))
\end{aligned}
\label{equation6}
\end{equation}
where $\alpha, \beta$ control the importance of the three terms.

\section{Experimental results}

\begin{figure}
	\begin{center}
		\includegraphics[width=\linewidth]{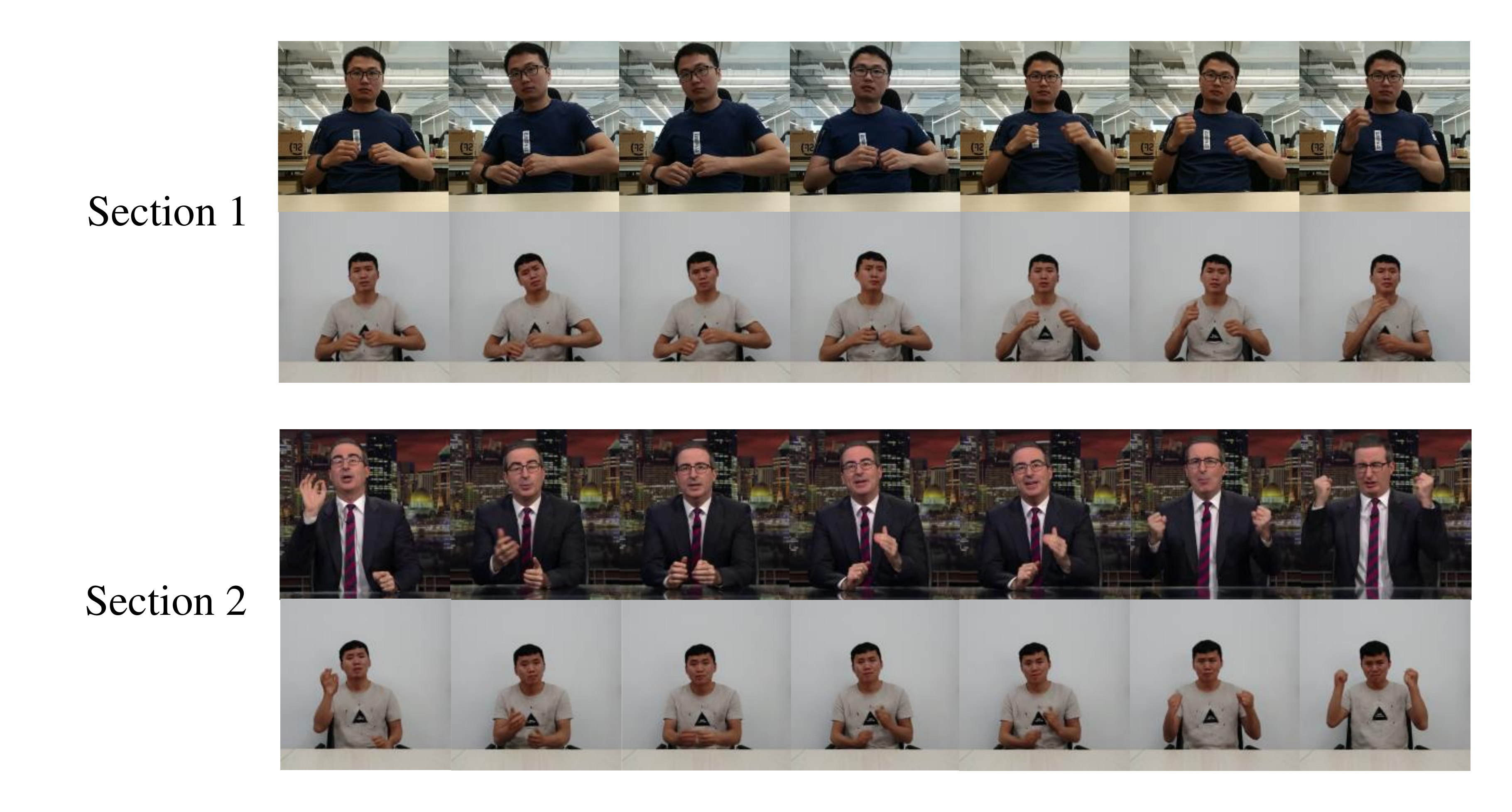}
	\end{center}
	\caption{Transfer results. In each section, the first row shows the source sequence, the second row shows synthesized target sequence.}
	\label{fig8}
\end{figure}

We conduct an experiment to synthesize two videos of one individual which are driven by two different individuals. Experimental results are shown in Fig.~\ref{fig8}. We can see that the synthesized target sequences are photorealistic and consistent with the source sequence in body motions, face expressions and poses, which is also a strong demonstration of our method's performance.

\section{Conclusion}

This paper presents a novel pipeline to perform human upper body video synthesis. Given an upper body video of a source person, we use upper body keypoints, facial action units and face poses as compact media to generate a target video consistent with the motion, expression, and pose of the source video. Experimental results show the effectiveness of our method. However, our method suffers from the problem of efficiency. Our future work will focus on exploring more lightweight networks to speed up the video synthesis.

\bibliographystyle{abbrv-doi}

\bibliography{template}
\end{document}